\documentclass[11pt]{article}

\usepackage[preprint]{arr_style/acl}

\usepackage{times}
\usepackage{latexsym}

\usepackage[T1]{fontenc}
\usepackage[utf8]{inputenc}
\usepackage{microtype}
\usepackage{inconsolata}
\usepackage{float}
\usepackage{adjustbox}
\usepackage{amsmath}
\usepackage{graphicx}
\usepackage[dvipsnames]{xcolor}
\usepackage{pifont}

\usepackage{graphicx}
\usepackage{wrapfig}
\usepackage{booktabs}
\usepackage{multirow}
\usepackage[table]{xcolor}
\usepackage{array}
\usepackage{amsmath}
\usepackage{amssymb}
\usepackage{pifont}
\usepackage{hyperref}
\usepackage{xspace}

\definecolor{darkgreen}{rgb}{0.0,0.5,0.0}

\usepackage{pifont}
\usepackage[table,xcdraw]{xcolor}
\usepackage[framemethod=TikZ]{mdframed}
\usepackage[most]{tcolorbox}
\newmdenv[
  backgroundcolor=gray!10,
  linecolor=black!40,
  topline=true,
  bottomline=true,
  rightline=true,
  leftline=true,
  linewidth=0.8pt,
  roundcorner=4pt,
  skipabove=10pt,
  skipbelow=10pt,
]{theobox}

\newtcolorbox{theoremBox}{
  colback=gray!10,
  colframe=black!40,
  arc=2pt,
  boxrule=0.7pt,
  left=6pt,
  right=6pt,
  top=6pt,
  bottom=6pt
}

\definecolor{myblue}{HTML}{215E9A}
\definecolor{darkgreen}{HTML}{006400}

\title{Temporal Gains, Spatial Costs: Revisiting Video Fine-Tuning in\\ Multimodal Large Language Models} 

\author{
\textbf{Linghao Zhang}\textsuperscript{1,4}\thanks{Equal contribution.}~
\textbf{Jungang Li}\textsuperscript{2,3}\footnotemark[1]~
\textbf{Yonghua Hei}\textsuperscript{2,3}~
\textbf{Sicheng Tao}\textsuperscript{2}~
\textbf{Song Dai}\textsuperscript{2,3}~\\
\textbf{Yibo Yan}\textsuperscript{2,3}~
\textbf{Zihao Dongfang}\textsuperscript{2,3}~
\textbf{Weiting Liu}\textsuperscript{5}~
\textbf{Chenxi Qin}\textsuperscript{7}~
\textbf{Hanqian Li}\textsuperscript{2}~\\
\textbf{Xin Zou}\textsuperscript{2,3}~
\textbf{Jiahao Zhang}\textsuperscript{2}~
\textbf{Shuhang Xun}\textsuperscript{6}~\\
\textbf{Haiyun Jiang}\textsuperscript{1}~\thanks{Corresponding author.}
\textbf{Xuming Hu}\textsuperscript{2,3}~\footnotemark[2]
\\[0.5em]
\textsuperscript{1}SJTU,
\textsuperscript{2}HKUST(GZ),
\textsuperscript{3}HKUST,
\textsuperscript{4}CityU,
\textsuperscript{5}FDU,
\textsuperscript{6}HIT,
\textsuperscript{7}TJU
}
\begin{document}

\maketitle

\begin{abstract}

Multimodal large language models (MLLMs) are typically trained in multiple stages, with video-based supervised fine-tuning (Video-SFT) serving as a key step for improving visual understanding. Yet its effect on the fine-grained evolution of visual capabilities, particularly the balance between spatial and temporal understanding, remains poorly understood. In this paper, we systematically study how Video-SFT reshapes visual capabilities in MLLMs. Across architectures, parameter scales, and frame sampling settings, we observe a consistent pattern: \textbf{Video-SFT reliably improves video performance, but often yields limited gains or even degradation on static image benchmarks.}
We further show that this trade-off is closely tied to temporal budget: increasing the number of sampled frames generally improves video performance, but does not reliably improve static image performance. Motivated by this finding, we study an instruction-aware Hybrid-Frame strategy that adaptively allocates frame counts and partially mitigates the image-video trade-off. Our results indicate that \textbf{Video-SFT is not a free lunch for MLLMs, and preserving spatial understanding remains a central challenge in joint image-video training.}
\end{abstract}

\section{Introduction}
\label{sec:intro}

\begin{figure}[h]
  \centering
\includegraphics[width=\linewidth]{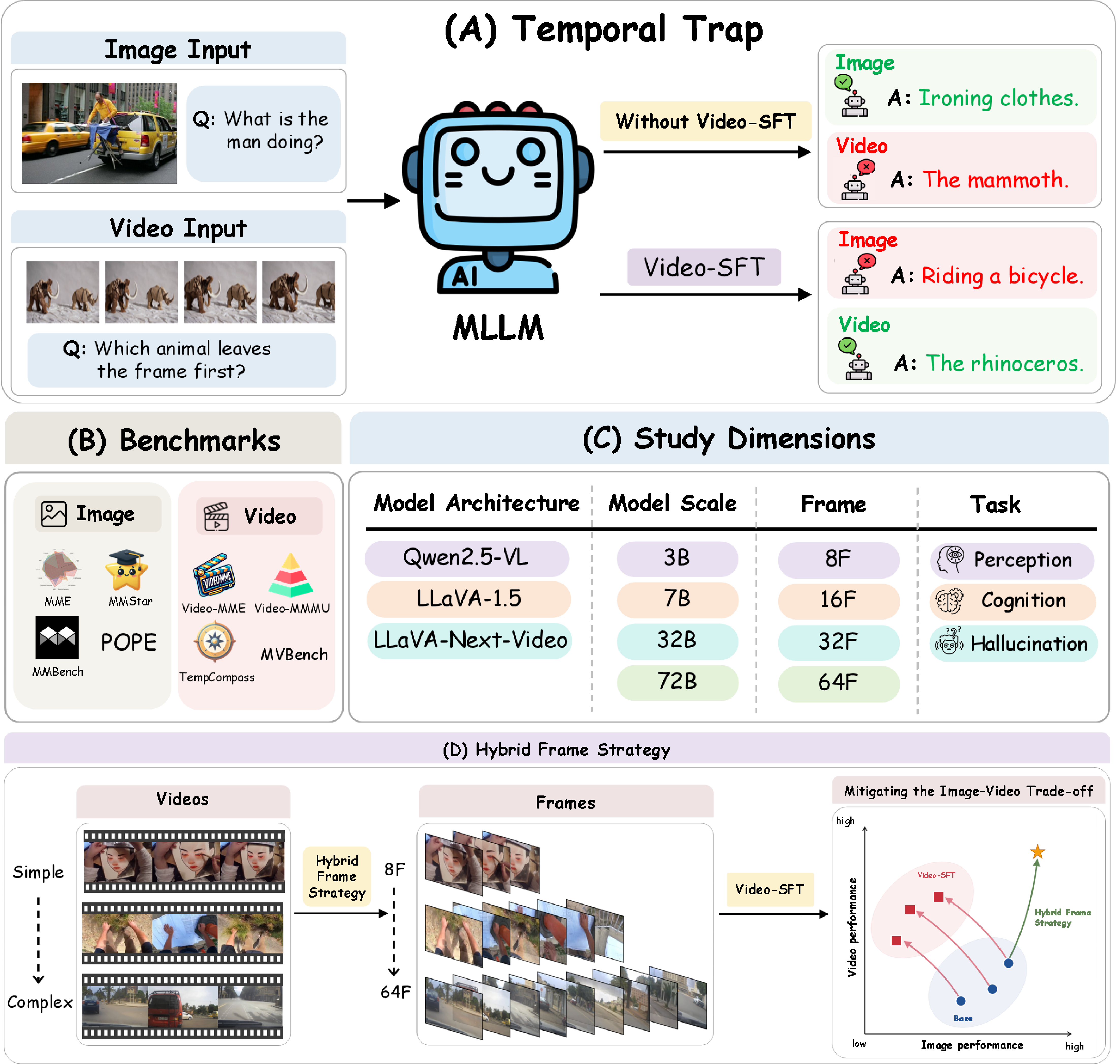}

    \caption{\textbf{Overview of the temporal trap and our study.}
\textbf{(A)} Illustration of the \emph{temporal trap}: Video-SFT improves video performance but can weaken spatial capability on static images.
\textbf{(B)} Image and video benchmarks used in our experiments.
\textbf{(C)} Main study dimensions: architecture, scale, frame budget, and task.
\textbf{(D)} \textbf{Hybrid-Frame Strategy}, which adaptively assigns a suitable frame budget to each training sample and partially mitigates the image--video trade-off.}
  
  \label{fig:1_illustration}

\end{figure}

The rapid progress of Multimodal Large Language Models (MLLMs) has substantially advanced visual understanding, extending model capabilities from static images to more general visual modeling over both images and videos~\cite{yin2024survey,xu2025lvlmehub,liu2025javisgpt,xun2025rtv}. 
Recent models such as Qwen2.5-VL~\cite{bai2025qwen2} and LLaVA-OneVision~\cite{li2024llava} show that unified language--vision frameworks can achieve strong performance across diverse tasks, including image captioning, visual question answering, and video reasoning~\cite{liu2023visual,dai2023instructblip}. 
Gemini 2.5~\cite{comanici2025gemini25pushingfrontier} utilizes a natively multimodal architecture to support long-context understanding, enabling the processing of up to 3 hours of video content. Kimi K2.5~\cite{kimiteam2026kimik25visualagentic} leverages joint text-vision pre-training and the MoonViT-3D architecture to enhance the understanding capabilities for both images and videos.

As videos can be naturally viewed as sequences of images, a growing line of work seeks to model image and video inputs within a shared visual space using common visual encoders and unified alignment mechanisms~\cite{jin2024chat,wang2023all,panagopoulou2024xinstructblip,tangunivit,zhang2024llavanextvideo}. 
Under this trend, video-based supervised fine-tuning (Video-SFT) has become a widely adopted post-training strategy for improving video understanding.

A common underlying assumption is that Video-SFT not only strengthens temporal modeling, but also benefits unified visual learning more broadly. 
If this assumption holds, then improving video understanding should at least preserve, if not enhance, the model's capability on static image tasks. 
However, despite the growing adoption of joint or staged image--video training, this assumption has not been systematically examined~\cite{gao2025tcllava,zeng2024foundation,li2025mvbench}. 
It remains unclear whether progress in video understanding reliably transfers to image understanding in MLLMs.

To study this question, we conduct a systematic analysis of how Video-SFT reshapes visual capabilities in MLLMs. Under a unified Video-SFT pipeline, we evaluate representative model families across architectural designs, parameter scales, and frame sampling settings on a broad set of image and video benchmarks. As shown in Figure~\ref{fig:1_illustration}, a consistent pattern emerges: Video-SFT reliably improves video performance, but often yields limited gains or even degradation on image benchmarks. We term this recurring image--video trade-off the temporal trap.

We further show that this trade-off is closely tied to temporal budget: increasing the number of sampled frames generally improves video performance, but does not reliably improve image performance. To better understand this behavior, we provide a conservative theoretical analysis that identifies sufficient conditions under which video-oriented updates can interfere with image objectives under shared-parameter optimization, and explains why larger frame budgets can intensify this conflict. Motivated by these findings, we study an instruction-aware \textbf{Hybrid-Frame Strategy} that adaptively allocates frame counts according to the spatiotemporal demands of each instruction. Experiments show that it can partially mitigate the trade-off while reducing redundant temporal exposure.

Our contributions are three-fold:
\begin{itemize}
\item[\ding{182}] We systematically study how Video-SFT reshapes image and video capabilities in MLLMs.
\item[\ding{183}] We identify a consistent image--video trade-off under Video-SFT, termed the \textbf{temporal trap}, and relate it to temporal budget.
\item[\ding{184}] We provide a conservative theoretical account of this trade-off and show that adaptive frame allocation can partially mitigate it.
\end{itemize}

\section{Related Work}
\label{sec: 2_related_work}
\subsection{Evolution of MLLMs}
\label{sec: 2_1_evolution_of_MLLMs}
Recent advancements in MLLMs have increasingly emphasized unified visual modeling, where images and videos are processed under a shared architectural and training framework
~\cite{huang2024image,shu2025audio}.  Qwen2.5-VL~\cite{bai2025qwen2} introduces Multimodal Rotary Position Embedding (MRoPE) to jointly encode spatial and temporal positions for tokens. Qwen3-VL~\cite{bai2025qwen3vltechnicalreport} further adopts an Interleaved-MRoPE, achieving full-frequency coverage of spatial–temporal information. 
Cambrian-1 investigates the role of visual tokens in image-centric MLLMs during both training and inference, while Cambrian-S extends this line to long-video spatial reasoning~\cite{tong2024cambrian,yang2025cambrian}. However, a systematic analysis of how Video-SFT influences unified cross-modal visual representation remains lacking.

\subsection{Challenges in Post-training of MLLMs}
\label{sec: 2_2_challenges_in_Post-training_of_MLLMs}
Recent studies show that continual tuning~\cite{shi2025continual} can lead to gradient conflicts~\cite{wei2025boosting} when models adapt to new tasks or modalities, introducing negative transfer and catastrophic forgetting~\cite{zhai2024investigating,lin2025continual,hua2025vision}. Recent benchmarks and frameworks have also focused on these challenges~\cite{yu2025progressive,zhao2025mllm}. 

Prior studies focus on modality conflicts between text and vision in MLLMs under instruction tuning, while conflicts between image and video modalities remain underexplored. In contrast, our work systematically investigates the balance between image and video capabilities in MLLMs under Video-SFT, and shows that adaptive frame allocation can partially mitigate the trade-off between image and video performance.
\section{Experimental Setup and Overview}
\label{3_experimental_setup_and_overview}

\subsection{Problem Setting}
\label{3_1_problem_setting}
Our study focuses on MLLMs under Video-SFT and systematically analyzes how Video-SFT affects two core visual capabilities: image understanding and video understanding. 
As videos can be naturally viewed as sequences of images, improvements in image understanding during the Video-SFT stage are expected.
However, our results reveal a different phenomenon: MLLMs exhibit a conflict between image and video modalities. After Video-SFT, video understanding improves while image understanding degrades. We refer to this phenomenon as the temporal trap.

\subsection{Study Dimensions}
\label{3_2_study_dimensions}
We conduct systematic experiments along three key dimensions. 
\begin{itemize}
\item[\ding{168}] \textbf{Model architecture}: Representative MLLMs including Qwen2.5-VL, LLaVA-Next-Video, and LLaVA-1.5. 
\item[\ding{169}] \textbf{Model scale}: Four scales of Qwen2.5-VL with 3B, 7B, 32B, and 72B parameters. 
\item[\ding{171}] \textbf{Frame sampling setting}: Videos are uniformly sampled with 8, 16, 32, and 64 frames during Video-SFT.
\end{itemize}

\subsection{Datasets and Evaluation}
\label{3_3_datasets_and_evaluations}
Based on the LLaVA-Next-Video-178k~\cite{zhang2024videoinstructiontuningsynthetic} dataset, we curate a training dataset of 20,000 videos collected from 10 different sources, covering diverse instruction formats such as textual descriptions, open-ended questions, and multiple-choice questions to ensure substantial data diversity. Training data statistics are reported in the supplementary material.

Evaluation datasets are selected from commonly used benchmarks for MLLMs. The image benchmarks include MME~\cite{fu2023mme}, MMStar~\cite{chen2024we}, MMBench~\cite{liu2024mmbench}, and POPE~\cite{Li-hallucination-2023}, while the video benchmarks consist of Video-MME~\cite{fu2025video}, MVBench~\cite{li2025mvbench}, TempCompass~\cite{liu2024tempcompass}, and Video-MMMU~\cite{hu2025video}. These benchmarks cover the core visual abilities, including coarse- and fine-grained perception, cognition, and hallucination.

\begin{figure}[h]
  \centering
  \adjincludegraphics[trim=0 0 {0.25\width} 0, clip, width=\linewidth]{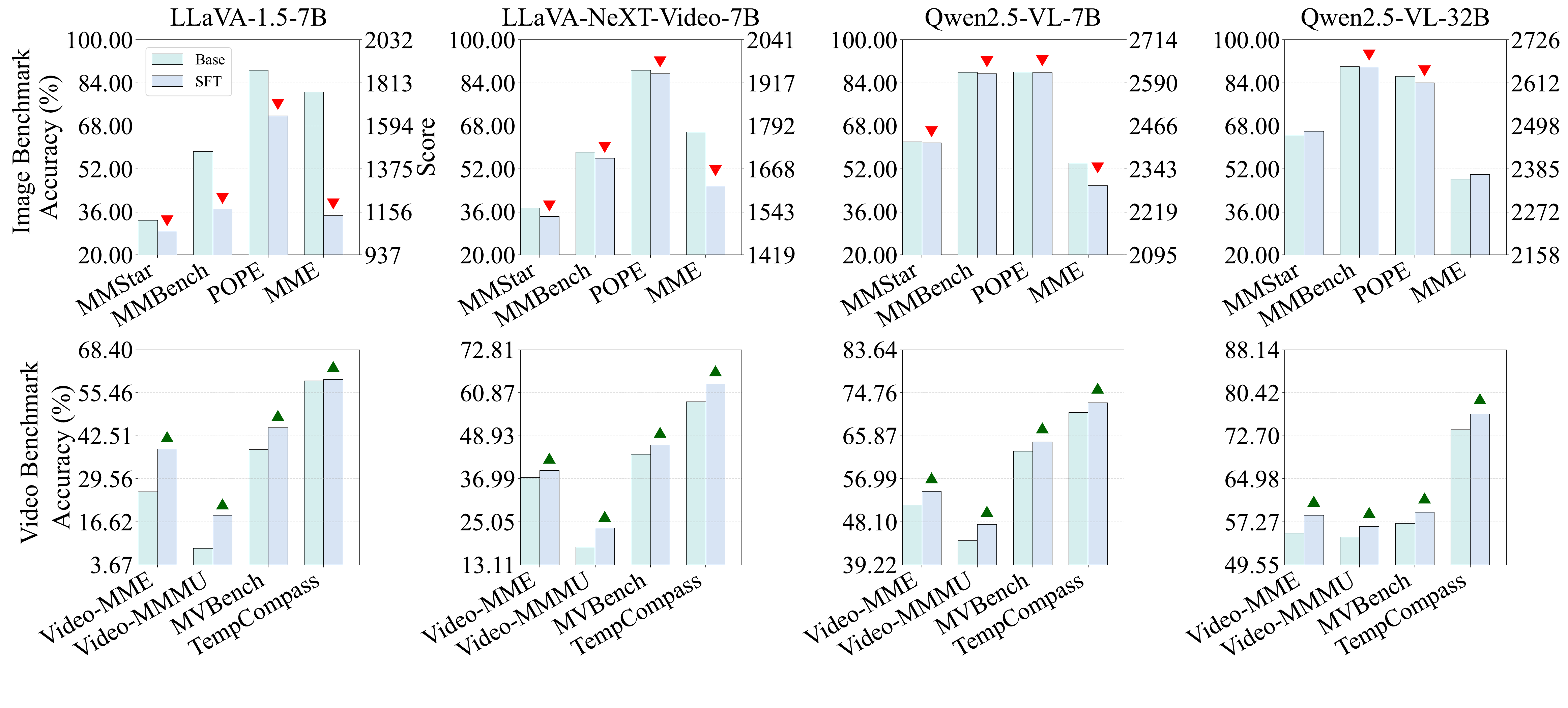}
    \caption{Comparison of image and video benchmark performance before and after Video-SFT across different MLLMs.
\textcolor{red}{$\blacktriangledown$} denotes performance degradation after SFT, whereas \textcolor{darkgreen}{$\blacktriangle$} indicates performance improvement.}
  
  \label{fig:3_base_results}
  \vspace{-8pt}
\end{figure}

\begin{figure}[b]
  \centering
  \includegraphics[
  width=1\linewidth,
]{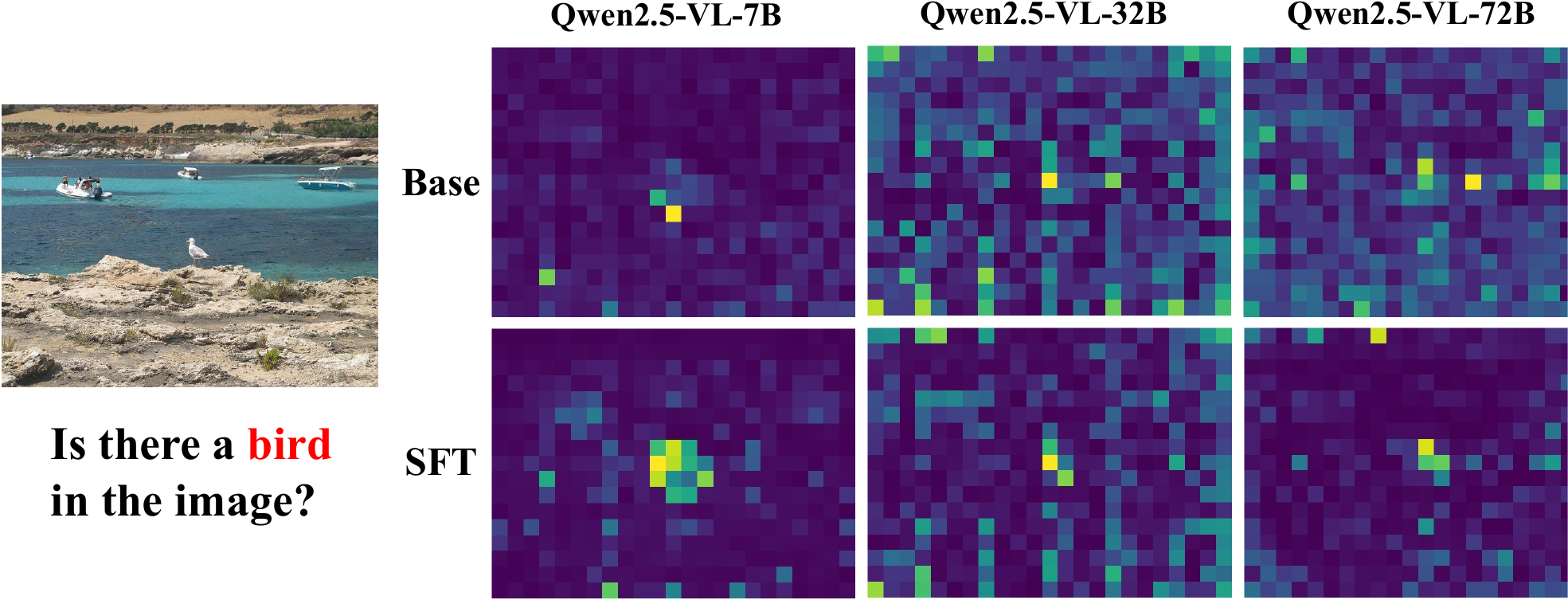}


    \caption{\textbf{Cross-scale attention visualizations before and after Video-SFT on Qwen2.5-VL models (7B, 32B, 72B).} For the query “Is there a bird in the image?”, attention becomes more dispersed in smaller models after Video-SFT, while larger models retain more localized focus on the target object, suggesting improved robustness to the temporal trap.}
  
  \label{fig:R_attation_map}

\end{figure}

\begin{figure*}[tbp]
  \centering
  \includegraphics[width=\linewidth]
  {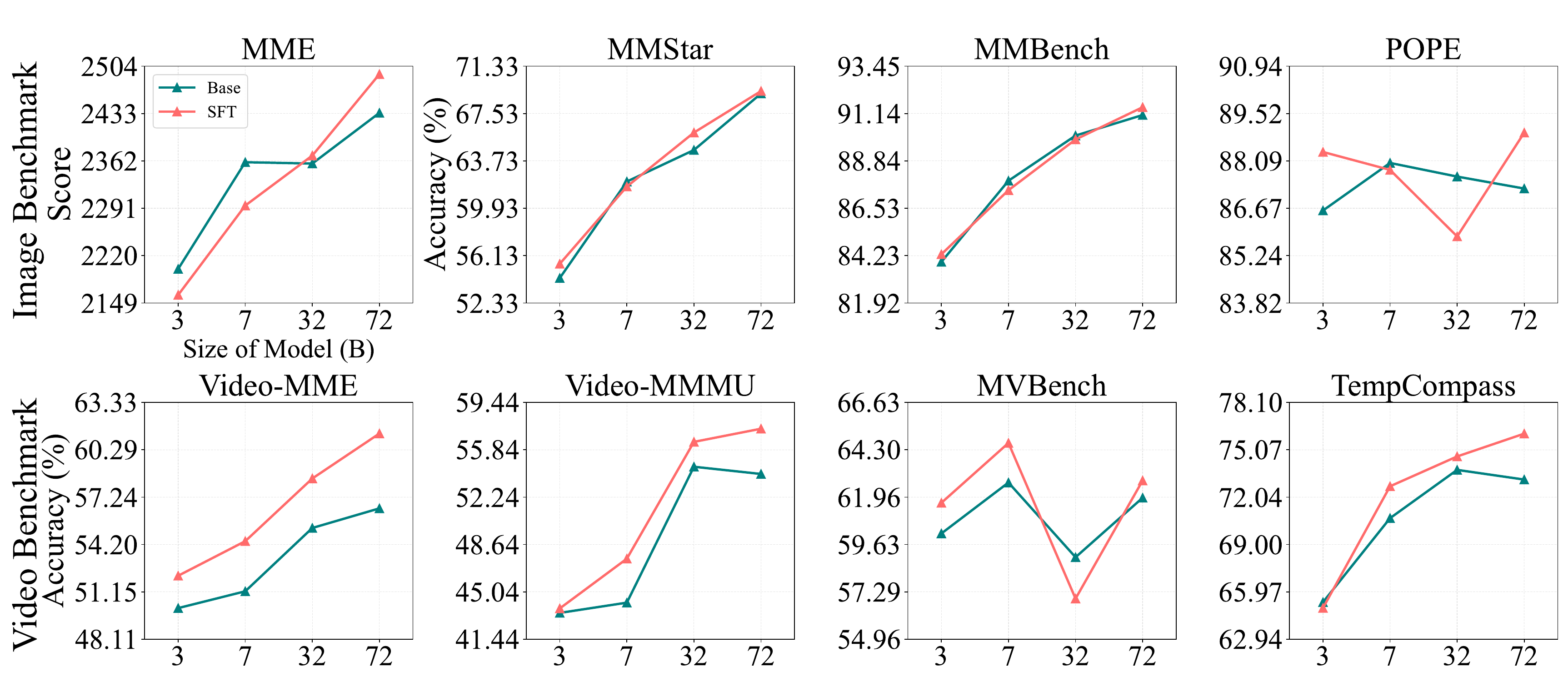}
    \caption{Comparison of image and video benchmark performance before and after Video-SFT across different scale Qwen2.5-VL models, including 3B, 7B, 32B and 72B parameters.}
  \label{fig:3_model_size}
\end{figure*}


\section{The Temporal Trap behind Visual Modality Conflict}

Although videos are composed of sequences of images and share the same visual encoder, improvements in video understanding do not transfer to static image understanding. We observe a systematic trade-off: Video-SFT enhances video performance while often degrading image performance. We refer to this phenomenon as the temporal trap, which reflects an intrinsic conflict between temporal adaptation and spatial visual reasoning. To better understand this phenomenon, we analyze it along three key dimensions: model architecture (Sec.~\ref{sec:3.1 model architecture}), model scale (Sec.~\ref{sec:3.2 model size}), and fine-tuning frame count (Sec.~\ref{sec:4.1 train frame}).

\subsection{Impact of Model Architecture}
\label{sec:3.1 model architecture}

Figure~\ref{fig:3_base_results} shows a consistent trend in all architectures evaluated. After Video-SFT, the performance on the video benchmarks improves for nearly every model, while most image benchmarks exhibit clear degradation. This pattern reveals a clear conflict between image and video modalities: although Video-SFT improves video understanding, it simultaneously weakens static image reasoning.

The magnitude of the conflict varies across architectures. LLaVA-1.5 exhibits the largest performance drop in the image benchmarks, while LLaVA-NeXT-Video shows a smaller gap. Qwen2.5-VL remains comparatively stable, indicating that stronger spatial–temporal alignment and mixed image–video pre-training can partially mitigate the conflict. Nevertheless, the temporal trap persists across all architectures.

\subsection{Impact of Model Size}
\label{sec:3.2 model size}
Figure~\ref{fig:3_model_size} shows that increasing model scale can partially mitigate the negative effect of Video-SFT on image understanding. However, this mitigation is not strictly monotonic. From 3B to 32B, the image benchmark performance after Video-SFT still exhibits noticeable fluctuations across datasets rather than a consistent improvement trend. For the 72B model, the post-SFT performance becomes comparable to, or even slightly better than, the base model on most image benchmarks.

Figure~\ref{fig:R_attation_map} provides further evidence for this observation. As the model size increases, the attention of the target object shifts from being scattered to being concentrated after Video-SFT. This suggests that larger models are better able to preserve stable spatial representations under the temporal trap.

Although the 72B model shows the most stable behavior, performance of models from 3B to 32B still shows fluctuations. Moreover, in many scenarios, the additional cost associated with using larger models is often prohibitive.

\subsection{Impact of Fine-tuning Frame Count}
\label{sec:4.1 train frame}

\begin{figure*}[tbp]
  \centering
  \includegraphics[width=\linewidth]
  {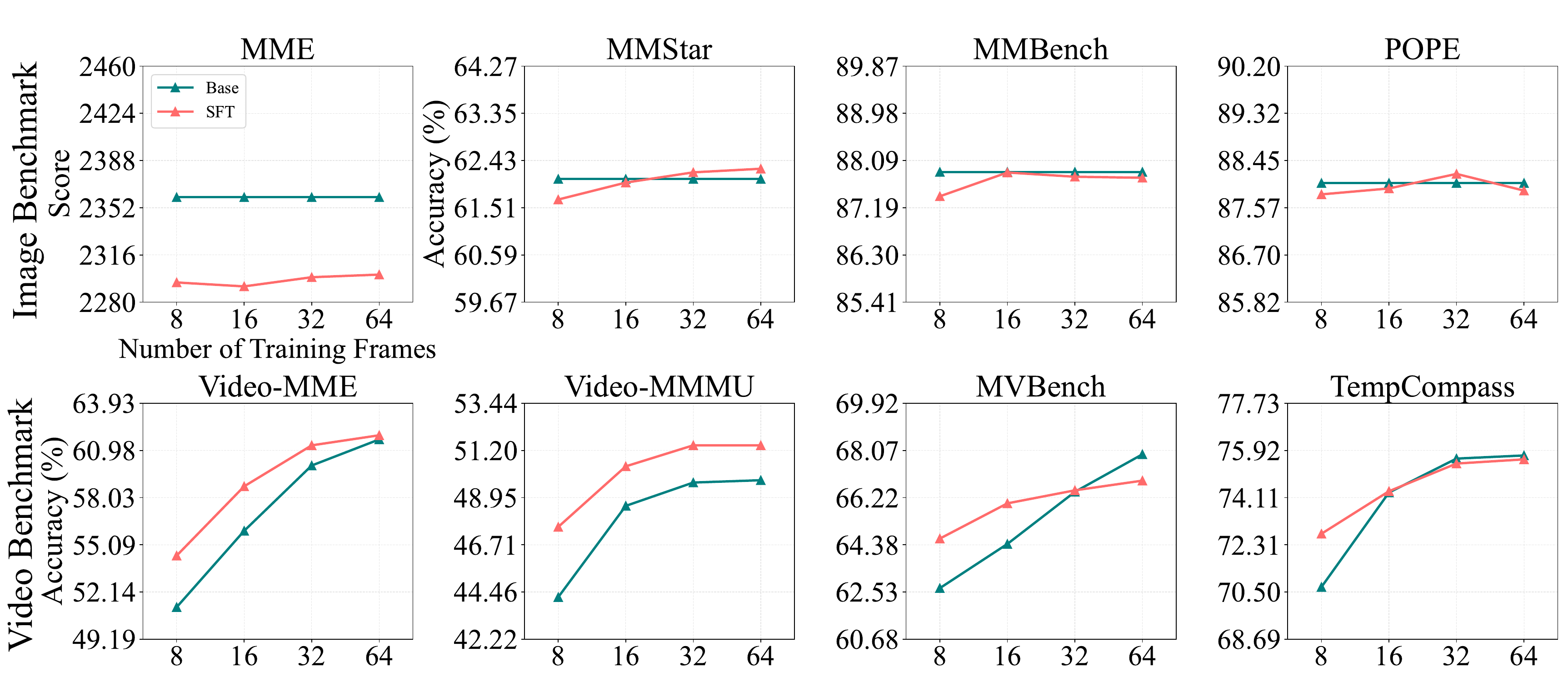}
    \caption{Comparison of performance on image and video benchmarks before and after Video-SFT across 8/16/32/64  training frames on Qwen2.5-VL-7B model.
}
  \label{fig:3_train_frame}
\end{figure*}

As shown in Figure~\ref{fig:3_train_frame}, increasing the number of training frames consistently improves the performance of the video benchmarks, confirming the importance of temporal information for video understanding. However, the gain gradually saturates as the frame count increases, indicating diminishing returns from additional temporal input.

For image benchmarks, performance on MME consistently underperforms the base model across all frame settings after Video-SFT. MMStar exhibits a gradual improvement as the number of training frames increases, but the gain clearly slows down at higher frame counts. The performance on MMBench and POPE exhibits an increase–then–decrease trend as the number of training frames increases.

These results suggest that redundant temporal information during Video-SFT can disrupt the model's static visual representations and weaken its generalization on image tasks, leading to the temporal trap phenomenon.

\section{Theoretical Analysis}
\label{sec:theory}

In this section, we provide a conservative theoretical account of why Video-SFT may improve video performance while degrading spatial capability in unified MLLMs, and why adaptive frame allocation can mitigate this effect. Rather than claiming a complete internal mechanism, we derive local sufficient conditions under which the observed image--video trade-off can arise under shared-parameter optimization.

\subsection{Preliminaries and Notation}

Let $\theta \in \mathbb{R}^d$ denote the trainable parameters of a unified MLLM. Let $x$ denote an image input, $v$ a video input, $q$ a textual instruction or question, and $y$ the supervision target (e.g., an answer token sequence or class label). Let $\ell(\cdot,\cdot)$ denote the training loss, and let $\Phi_m(v)$ be a frame sampling operator that extracts $m$ frames from video $v$, where $m$ is the temporal budget.

We consider the population objectives
\begin{equation}
\mathcal{L}_{\mathrm{img}}(\theta)
=
\mathbb{E}_{(x,q,y)\sim \mathcal{D}_{\mathrm{img}}}
\big[
\ell(f_\theta(x,q),y)
\big],
\label{eq:limg}
\end{equation}
and
\begin{equation}
\scalebox{0.83}{$
\mathcal{L}_{\mathrm{vid}}^{(m)}(\theta)
=
\mathbb{E}_{(v,q,y)\sim \mathcal{D}_{\mathrm{vid}}}
\big[
\ell(f_\theta(\Phi_m(v),q),y)
\big],
$}
\label{eq:lvid}
\end{equation}
where the superscript $m$ emphasizes that the video objective depends on the temporal budget.

We write
\begin{equation}
g_{\mathrm{img}}
:=
\nabla_\theta \mathcal{L}_{\mathrm{img}}(\theta),
g_{\mathrm{vid}}^{(m)}
:=
\nabla_\theta \mathcal{L}_{\mathrm{vid}}^{(m)}(\theta),
\label{eq:grads}
\end{equation}
and denote the corresponding Hessians by
\begin{equation}
\scalebox{0.83}{$
H_{\mathrm{img}}(\theta)
:=
\nabla_\theta^2 \mathcal{L}_{\mathrm{img}}(\theta),
H_{\mathrm{vid}}^{(m)}(\theta)
:=
\nabla_\theta^2 \mathcal{L}_{\mathrm{vid}}^{(m)}(\theta).
$}
\label{eq:hessians}
\end{equation}

A single Video-SFT gradient step updates parameters as
\begin{equation}
\theta^+ = \theta - \eta g_{\mathrm{vid}}^{(m)},
\label{eq:update}
\end{equation}
where $\eta > 0$ is the learning rate.

\paragraph{Definition 1 (Gradient alignment).}
For two objectives $\mathcal{L}_a$ and $\mathcal{L}_b$, define their local alignment at $\theta$ as
\begin{equation}
\mathrm{Align}(\mathcal{L}_a,\mathcal{L}_b;\theta)
:=
\left\langle
\nabla_\theta \mathcal{L}_a(\theta),
\nabla_\theta \mathcal{L}_b(\theta)
\right\rangle.
\label{eq:alignment}
\end{equation}
Positive alignment indicates locally cooperative optimization directions, while negative alignment indicates local conflict.

\paragraph{Assumption 1 (Local smoothness).}
There exist constants $\beta_{\mathrm{img}}>0$ and $\beta_{\mathrm{vid}}^{(m)}>0$ such that $\nabla_\theta \mathcal{L}_{\mathrm{img}}$ and $\nabla_\theta \mathcal{L}_{\mathrm{vid}}^{(m)}$ are Lipschitz continuous in a neighborhood of $\theta$. Equivalently, whenever the Hessians exist in that neighborhood,
\begin{equation}
\scalebox{0.83}{$
\|H_{\mathrm{img}}(\theta')\|_2 \le \beta_{\mathrm{img}},
\quad
\|H_{\mathrm{vid}}^{(m)}(\theta')\|_2 \le \beta_{\mathrm{vid}}^{(m)}
$}
\end{equation}
for all $\theta'$ in that neighborhood.

\paragraph{Assumption 2 (Shared-parameter coupling).}
Image and video objectives are optimized through the same parameter vector $\theta$, so updates for one objective can affect the other through gradient interaction.

\subsection{A First-Order Condition for Image Degradation Under Video-SFT}

Video-SFT directly optimizes $\mathcal{L}_{\mathrm{vid}}^{(m)}$, not $\mathcal{L}_{\mathrm{img}}$. Whether spatial capability is preserved therefore depends on how the video gradient aligns with the image gradient in the shared parameter space.

By the second-order Taylor theorem, there exists a point
$\tilde{\theta}_{\mathrm{img}}$ on the line segment between $\theta$ and $\theta^+$ such that
\begin{equation}
\begin{aligned}
\mathcal{L}_{\mathrm{img}}(\theta^+)
&=
\mathcal{L}_{\mathrm{img}}(\theta)
-
\eta \langle g_{\mathrm{img}}, g_{\mathrm{vid}}^{(m)} \rangle \\
&\quad
+
\frac{\eta^2}{2}
\big(g_{\mathrm{vid}}^{(m)}\big)^\top
H_{\mathrm{img}}(\tilde{\theta}_{\mathrm{img}})
g_{\mathrm{vid}}^{(m)} .
\end{aligned}
\label{eq:taylor_img}
\end{equation}

\paragraph{Proposition 1 (Local sufficient condition for image loss increase).}
Assume Assumption 1 holds and $g_{\mathrm{vid}}^{(m)} \neq 0$. If
\begin{equation}
\langle g_{\mathrm{img}}, g_{\mathrm{vid}}^{(m)} \rangle < 0,
\label{eq:conflict_condition}
\end{equation}
then there exists $\eta_0>0$ such that, for all $0<\eta<\eta_0$, one Video-SFT step increases the image loss:
\begin{equation}
\mathcal{L}_{\mathrm{img}}(\theta^+) > \mathcal{L}_{\mathrm{img}}(\theta).
\label{eq:img_increase}
\end{equation}
In particular, it suffices to take
\begin{equation}
0 < \eta <
\frac{
-2\langle g_{\mathrm{img}}, g_{\mathrm{vid}}^{(m)} \rangle
}{
\beta_{\mathrm{img}}\|g_{\mathrm{vid}}^{(m)}\|_2^2
}.
\label{eq:eta_condition}
\end{equation}

\paragraph{Proof.}
From Eq.~(\ref{eq:taylor_img}) and Assumption 1,
\begin{equation}
\left|
\big(g_{\mathrm{vid}}^{(m)}\big)^\top
H_{\mathrm{img}}(\tilde{\theta}_{\mathrm{img}})
g_{\mathrm{vid}}^{(m)}
\right|
\le
\beta_{\mathrm{img}}
\|g_{\mathrm{vid}}^{(m)}\|_2^2.
\label{eq:rayleigh}
\end{equation}
Therefore,
\begin{equation}
\begin{aligned}
\mathcal{L}_{\mathrm{img}}(\theta^+) - \mathcal{L}_{\mathrm{img}}(\theta)
&\ge
-\eta \langle g_{\mathrm{img}}, g_{\mathrm{vid}}^{(m)} \rangle \\
&\quad
-\frac{\beta_{\mathrm{img}}}{2}\eta^2
\|g_{\mathrm{vid}}^{(m)}\|_2^2 .
\end{aligned}
\label{eq:img_bound}
\end{equation}
If Eq.~(\ref{eq:conflict_condition}) holds, then the first term on the right-hand side is strictly positive. Moreover, if Eq.~(\ref{eq:eta_condition}) holds, the right-hand side of Eq.~(\ref{eq:img_bound}) remains strictly positive. Hence $\mathcal{L}_{\mathrm{img}}(\theta^+) > \mathcal{L}_{\mathrm{img}}(\theta)$. \hfill $\square$

Proposition~1 formalizes a local sufficient condition under which Video-SFT can be harmful to image performance: a video-driven update may still increase the image loss when the two objectives are negatively aligned in the shared parameter space. This is a standard negative-transfer phenomenon in shared-parameter learning~\cite{yu2020gradient,zhang2022survey}.

By the descent lemma under Assumption~1, the same update satisfies
\begin{equation}
\scalebox{0.83}{$
\mathcal{L}_{\mathrm{vid}}^{(m)}(\theta^+)
\le
\mathcal{L}_{\mathrm{vid}}^{(m)}(\theta)
-
\eta
\left(
1-\frac{\beta_{\mathrm{vid}}^{(m)}}{2}\eta
\right)
\|g_{\mathrm{vid}}^{(m)}\|_2^2 .
$}
\label{eq:taylor_vid}
\end{equation}
Hence, for any $0<\eta<2/\beta_{\mathrm{vid}}^{(m)}$, the video objective decreases. Thus, video improvement and image degradation are not contradictory; they can coexist when the video update is locally beneficial for $\mathcal{L}_{\mathrm{vid}}^{(m)}$ but negatively aligned with $\mathcal{L}_{\mathrm{img}}$.

\paragraph{Remark 1 (Population objective and minibatch training).}
Proposition~1 is a local population-level statement. In practical Video-SFT, the full gradient $g_{\mathrm{vid}}^{(m)}$ is replaced by a minibatch estimator. Under standard unbiasedness assumptions, the same result motivates the expected tendency of image loss increase when the expected alignment is negative.

\paragraph{Remark 2 (Implication for multi-stage post-training).}
In current MLLM pipelines, Video-SFT is typically applied as a late-stage post-training phase starting from a checkpoint that already has strong spatial capability. A smaller learning rate reduces the magnitude of each individual update, but repeated small updates with persistently biased alignment can still accumulate into measurable spatial degradation over training, especially because the image objective is no longer explicitly optimized in this phase.

\subsection{Temporal Budget as a Source of Gradient Bias}

The previous result explains \emph{when} image degradation can happen. We now analyze why the temporal budget $m$ can affect its severity.

For analytical convenience, we consider the following stylized local decomposition:
\begin{equation}
g_{\mathrm{vid}}^{(m)}
=
g_{\mathrm{sh}}
+
\alpha(m)\, g_{\mathrm{tmp}}
+
\varepsilon^{(m)},
\label{eq:gradient_decomp}
\end{equation}
where $g_{\mathrm{sh}}$ denotes a shared visual component useful to both image and video understanding, $g_{\mathrm{tmp}}$ denotes a temporally specialized component induced by video-specific adaptation, and $\varepsilon^{(m)}$ is a residual term capturing sampling noise, redundancy, and sample-specific nuisance variation. The coefficient $\alpha(m)\ge 0$ measures how strongly temporal specialization enters the update as more frames are used.

Taking inner products with $g_{\mathrm{img}}$ gives
\begin{equation}
\scalebox{0.83}{$
\begin{aligned}
\langle g_{\mathrm{img}}, g_{\mathrm{vid}}^{(m)} \rangle
&=
\langle g_{\mathrm{img}}, g_{\mathrm{sh}} \rangle
+
\alpha(m)\langle g_{\mathrm{img}}, g_{\mathrm{tmp}} \rangle \\
&\quad
+
\langle g_{\mathrm{img}}, \varepsilon^{(m)} \rangle.
\end{aligned}
$}
\label{eq:alignment_decomp}
\end{equation}

We consider the following average-case assumptions.

\paragraph{Assumption 3 (Positive shared alignment).}
\begin{equation}
\mathbb{E}\big[\langle g_{\mathrm{img}}, g_{\mathrm{sh}} \rangle\big] > 0.
\end{equation}
This reflects the fact that image and video tasks share nontrivial spatial semantics.

\paragraph{Assumption 4 (Non-positive temporal alignment).}
\begin{equation}
\mathbb{E}\big[\langle g_{\mathrm{img}}, g_{\mathrm{tmp}} \rangle\big] \le 0.
\end{equation}
This captures the possibility that temporally specialized adaptation competes with spatial capability preservation in shared parameters.

\paragraph{Assumption 5 (Unbiased residual interaction).}
\begin{equation}
\mathbb{E}\big[\langle g_{\mathrm{img}}, \varepsilon^{(m)} \rangle\big] = 0,
\end{equation}
while $\mathrm{Var}(\varepsilon^{(m)})$ is non-decreasing in $m$ once additional frames become redundant for a subset of samples.

\begin{equation}
\mathbb{E}\!\left[\left\langle g_{\mathrm{img}}, g_{\mathrm{vid}}^{(m)} \right\rangle\right]
= \rho_{\mathrm{sh}} - \alpha(m)\rho_{\mathrm{tmp}} .
\label{eq:expected_alignment}
\end{equation}
where
\begin{equation}
\begin{aligned}
\rho_{\mathrm{sh}}
&:= \mathbb{E}\!\left[\left\langle g_{\mathrm{img}}, g_{\mathrm{sh}} \right\rangle\right] > 0, \\
\rho_{\mathrm{tmp}}
&:= -\,\mathbb{E}\!\left[\left\langle g_{\mathrm{img}}, g_{\mathrm{tmp}} \right\rangle\right] \ge 0 .
\end{aligned}
\label{eq:rhos}
\end{equation}

Let $\mathcal{M}\subset \mathbb{N}$ denote the set of admissible frame budgets.

\paragraph{Proposition 2 (A discrete temporal-budget threshold).}
Suppose $\alpha(m)$ is non-decreasing on $\mathcal{M}$, and define
\begin{equation}
m^\star
:=
\min
\big\{
m\in \mathcal{M}:
\alpha(m)\rho_{\mathrm{tmp}} \ge \rho_{\mathrm{sh}}
\big\},
\label{eq:threshold}
\end{equation}
provided the set is nonempty. Then
\begin{equation}
\mathbb{E}\big[\langle g_{\mathrm{img}}, g_{\mathrm{vid}}^{(m)} \rangle\big]
\begin{cases}
> 0, & m < m^\star,\\
\le 0, & m \ge m^\star.
\end{cases}
\label{eq:threshold_result}
\end{equation}
Moreover, if for some admissible $m$ one has
$\alpha(m)\rho_{\mathrm{tmp}} = \rho_{\mathrm{sh}}$, then the expected alignment is exactly zero at that budget.

\paragraph{Interpretation.}
Proposition~2 formalizes one sufficient mechanism by which increasing temporal budget can flip the expected transfer from cooperative to conflicting. As $m$ grows, the update places increasing weight on temporally specialized adaptation. Once this component dominates the shared spatial benefit, the average alignment with the image objective becomes non-positive.

\subsection{Why Adaptive Frame Allocation is Theoretically Justified}

The previous results imply that temporal budget should not be treated as a globally fixed constant. We now formalize why sample-adaptive frame allocation is a sensible intervention.

Let $(V,Q,Y)$ denote the random video--instruction--target triple. Assume there exists a sample-wise minimal sufficient temporal budget $m_{\min}(V,Q)$ such that
\begin{equation}
Y \perp\!\!\!\perp V
\;|\;
\Phi_{m_{\min}(V,Q)}(V),\, Q.
\label{eq:min_sufficient}
\end{equation}
For a realized sample $(v,q)$, we write $m_{\min}(v,q)$ for its minimal sufficient budget. This assumption captures the fact that some instructions require only sparse temporal evidence, while others require denser temporal coverage.

Let $m^\star := m_{\min}(v,q)$. For any fixed $(v,q)$ and any budget
$m \ge m^\star$, if the additional frames beyond $m^\star$ are predominantly
redundant, they need not improve alignment with the image objective, but can
increase the second moment of the video gradient. We summarize this regime by
\begin{equation}
\mathbb{E}\!\left[
\left\langle g_{\mathrm{img}}, g_{\mathrm{vid}}^{(m)} \right\rangle
\right]
\le
\mathbb{E}\!\left[
\left\langle g_{\mathrm{img}}, g_{\mathrm{vid}}^{(m^\star)} \right\rangle
\right],
\label{eq:moment_condition_a}
\end{equation}
where both expectations are conditioned on the fixed pair $(v,q)$.

and
\begin{equation}
\scalebox{0.83}{$
\mathbb{E}\big[
\|g_{\mathrm{vid}}^{(m)}\|_2^2
\mid v,q
\big]
\ge
\mathbb{E}\big[
\|g_{\mathrm{vid}}^{(m_{\min}(v,q))}\|_2^2
\mid v,q
\big].
$}
\label{eq:moment_condition_b}
\end{equation}

By Assumption~1, the image objective satisfies the smoothness bound
\begin{equation}
\scalebox{0.83}{$
\begin{aligned}
\mathcal{L}_{\mathrm{img}}(\theta^+) - \mathcal{L}_{\mathrm{img}}(\theta)
\le\;&
-\eta \langle g_{\mathrm{img}}, g_{\mathrm{vid}}^{(m)} \rangle \\
&+
\frac{\beta_{\mathrm{img}}}{2}\eta^2
\|g_{\mathrm{vid}}^{(m)}\|_2^2 .
\end{aligned}
$}
\label{eq:img_upper}
\end{equation}

\paragraph{Proposition 3 (Adaptive budgeting under redundancy).}
Assume Eq.~(\ref{eq:min_sufficient}), Eq.~(\ref{eq:moment_condition_a}), and Eq.~(\ref{eq:moment_condition_b}) hold. Then for any realized sample $(v,q)$, choosing $m(v,q)=m_{\min}(v,q)$ minimizes the smoothness-based upper bound
\begin{equation}
\begin{aligned}
&-\eta \,\mathbb{E}\!\left[
\left\langle g_{\mathrm{img}}, g_{\mathrm{vid}}^{(m)} \right\rangle
\,\middle|\, v,q
\right] \\
&\quad+
\frac{\beta_{\mathrm{img}}}{2}\eta^2
\mathbb{E}\!\left[
\left\|g_{\mathrm{vid}}^{(m)}\right\|_2^2
\,\middle|\, v,q
\right]
\end{aligned}
\label{eq:prop3_bound}
\end{equation}
among all choices $m \ge m_{\min}(v,q)$.

\paragraph{Interpretation.}
Proposition~3 justifies adaptive frame allocation: use fewer frames for temporally simple samples and more for temporally demanding ones. When temporal sufficiency is sample-dependent, a sample-wise budget is better than a uniform one. Hybrid-Frame follows this principle by preserving necessary temporal evidence while avoiding redundancy.

\subsection{Summary and Connection to Empirical Findings}

The analysis suggests two main points. First, Video-SFT can improve video performance while degrading spatial capability when video-oriented updates are negatively aligned with the image objective. Second, this trade-off can intensify with temporal budget if larger frame counts strengthen temporally specialized updates more than shared spatial benefit. Thus, the observed image--video trade-off can arise naturally from shared-parameter optimization.

\paragraph{Connection to Empirical Findings.}
Proposition~1 accounts for the coexistence of video gains and spatial degradation. Proposition~2 shows how increasing frame budget can turn expected transfer from cooperative to conflicting. Proposition~3 motivates adaptive frame allocation as a conservative way to reduce redundant temporal exposure. Together, these results provide a principled lens on the observed temporal trap.
\section{Hybrid-Frame Strategy}
\label{sec:4.2 Hybrid-Frame Strategy}

Motivated by the adaptive budgeting principle in Eq.~(\ref{eq:min_sufficient}) and Proposition~3, we implement a \textbf{Hybrid-Frame Strategy} and evaluate it empirically. The goal is simple: allocate enough frames to preserve task-relevant temporal evidence, while avoiding redundant temporal exposure. We compare three frame allocation schemes: (i) a DINOv2-based strategy using inter-frame similarity, (ii) a VLM-based predictor built on Qwen2.5-VL-3B, and (iii) a VLM-based predictor built on Qwen3-VL-8B. As shown in Table~\ref{tab:compare_hybrid}, the DINOv2-based method is not sufficiently reliable, whereas both VLM-based predictors yield consistent improvements. Although Qwen3-VL-8B achieves the best overall performance, Qwen2.5-VL-3B also performs competitively, suggesting that instruction-aware frame allocation is effective even with relatively small predictors.


\begin{table}[h]
\centering
\scriptsize
\caption{
\textbf{Comparison of frame allocation strategies for Video-SFT on the Qwen2.5-VL-7B, training and inference using 8 frames.} 
}
\label{tab:compare_hybrid}

\setlength{\tabcolsep}{5pt}
\renewcommand{\arraystretch}{1.0}

\resizebox{\linewidth}{!}{
\begin{tabular}{lcccc}
\toprule
\multicolumn{1}{c}{\textbf{Setting}} 
& \multicolumn{2}{c}{\textbf{Image}} 
& \multicolumn{2}{c}{\textbf{Video}} \\
\cmidrule(lr){2-3} \cmidrule(lr){4-5}
& \textbf{MMStar} & \textbf{POPE} & \textbf{Video-MME} & \textbf{Video-MMMU} \\
\midrule

Base
& 62.07 & 88.03 & 51.19 & 44.22 \\

DINOv2-based
& 61.40{\textcolor{red}{\scriptsize$\downarrow$}}
& 87.91{\textcolor{red}{\scriptsize$\downarrow$}}
& 53.96{\textcolor{ForestGreen}{\scriptsize$\uparrow$}}
& 47.13{\textcolor{ForestGreen}{\scriptsize$\uparrow$}} \\

Qwen2.5-VL-3B-based
& \underline{62.19}{\textcolor{ForestGreen}{\scriptsize$\uparrow$}}
& \underline{88.19}{\textcolor{ForestGreen}{\scriptsize$\uparrow$}}
& \underline{53.47}{\textcolor{ForestGreen}{\scriptsize$\uparrow$}}
& \textbf{47.91}{\textcolor{ForestGreen}{\scriptsize$\uparrow$}} \\

Qwen3-VL-8B-based
& \textbf{62.33}{\textcolor{ForestGreen}{\scriptsize$\uparrow$}}
& \textbf{88.20}{\textcolor{ForestGreen}{\scriptsize$\uparrow$}}
& \textbf{54.96}{\textcolor{ForestGreen}{\scriptsize$\uparrow$}}
& \underline{47.22}{\textcolor{ForestGreen}{\scriptsize$\uparrow$}} \\

\bottomrule
\end{tabular}
}
\vspace{-4pt}
\end{table}

We then apply the Qwen3-VL-8B-based Hybrid-Frame Strategy to Video-SFT for Qwen2.5-VL-7B. As shown in Table~\ref{tab:total_model_frame}, Hybrid-Frame achieves the best accuracy on MMStar and POPE, outperforming models trained with larger fixed frame budgets such as 32 or 64 frames. At the same time, it maintains strong gains on video performance. This is consistent with the theoretical motivation in Eq.~(\ref{eq:prop3_bound}): once temporal sufficiency is reached, reducing redundant frames can improve the image--video trade-off without weakening useful supervision.

\begin{table}[h]
\centering
\small
\tiny
\setlength{\tabcolsep}{4pt}
\renewcommand{\arraystretch}{0.95}
\caption{
\textbf{Unified comparison of Video-SFT strategies across model architectures and frame budgets.} For Qwen2.5-VL-7B, training and inference frame counts follow the settings in parentheses, while LLaVA-1.5-7B are evaluated with a fixed 8-frame inference.
}
\label{tab:total_model_frame}
\vspace{2pt}
\resizebox{\linewidth}{!}{
\begin{tabular}{>{\raggedright\arraybackslash}l c c c c}
\toprule
\multicolumn{1}{c}{\textbf{Setting}} 
& \multicolumn{2}{c}{\textbf{Image}} 
& \multicolumn{2}{c}{\textbf{Video}} \\
\cmidrule(lr){2-3} \cmidrule(lr){4-5}
& \textbf{MMStar} & \textbf{POPE} & \textbf{Video-MME} & \textbf{Video-MMMU} \\
\midrule

\multicolumn{5}{l}{\textbf{Qwen2.5-VL-7B}} \\
\cmidrule(lr){1-5}
Video-SFT (8F)   
& 61.67 
& 87.82 
& 54.41
& \textbf{47.56} \\
\rowcolor[RGB]{245,248,255} Hybrid (8F)
& \textbf{62.33} 
& \textbf{88.20} 
& \textbf{54.96} 
& 47.22 \\
\cmidrule(lr){1-5}

Video-SFT (16F)  
& 62.00 
& 87.93 
& \textbf{58.74} 
& 50.44 \\
\rowcolor[RGB]{245,248,255} Hybrid (16F)
& \textbf{62.33} 
& \textbf{88.20} 
& 58.15
& \textbf{51.78}  \\
\cmidrule(lr){1-5}

Video-SFT (32F)  
& 62.20 
& \textbf{88.20} 
& 61.30
& 51.44 \\
\rowcolor[RGB]{245,248,255} Hybrid (32F)
& \textbf{62.33} 
& \textbf{88.20} 
& \textbf{61.52} 
& \textbf{55.12} \\
\cmidrule(lr){1-5}

Video-SFT (64F)  
& 62.27
& 87.89  
& \textbf{61.93} 
& 51.44  \\
\rowcolor[RGB]{245,248,255} Hybrid (64F)
& \textbf{62.33} 
& \textbf{88.20} 
& 60.93 
& \textbf{55.91} \\
\midrule

\multicolumn{5}{l}{\textbf{LLaVA-1.5-7B}} \\
\cmidrule(lr){1-5}
Video-SFT-8F
& 28.87 
& 71.69 
& 38.59 
& \underline{18.56} \\
Video-SFT-16F
& 28.47
& 72.78
& 51.39
& 16.44 \\
Video-SFT-32F
& \underline{29.40} 
& \underline{73.40} 
& \textbf{59.90} 
& 12.92  \\
\rowcolor[RGB]{245,248,255} \textbf{Hybrid}
& \textbf{32.47} 
& \textbf{75.66} 
& \underline{52.41} 
& \textbf{18.71} \\
\bottomrule
\end{tabular}
}

\vspace{-6pt}
\end{table}

These results show that Hybrid-Frame is an effective and practical intervention for mitigating the temporal trap. It improves the balance between spatial and temporal capability while also reducing unnecessary training cost.

To test whether this effect extends beyond the Qwen family, we further evaluate the same strategy on LLaVA-1.5-7B. As shown in Table~\ref{tab:total_model_frame}, Hybrid-Frame remains effective across different Video-SFT budgets, indicating that its benefit is not tied to a specific architecture. This supports the view that adaptive frame allocation is a broadly useful mechanism for reducing the image--video trade-off in unified MLLMs.

\section{Conclusion}

We systematically study how Video-SFT reshapes visual capabilities in MLLMs. Across architectures, scales, and frame budgets, a consistent pattern emerges: Video-SFT improves temporal understanding, but often weakens spatial capability. This suggests that current unified training pipelines have not yet achieved true image--video synergy. We further show that this trade-off is closely tied to temporal budget, and that adaptive frame allocation can partially mitigate it. Preserving spatial capability under Video-SFT remains a central challenge for joint image--video training.
\clearpage
\section*{Limitations}
\label{sec:7_limitations}
Our study covers representative MLLMs, but not the full space of architectures, training schemes, or evaluation settings. In particular, many recent MLLMs adopt streaming or online training, which may require different modeling and analysis. Our experiments also focus on standard image and video benchmarks, excluding settings such as streaming inputs and interactive multimodal reasoning. Finally, the proposed Hybrid-Frame strategy is heuristic rather than fully principled.

\bibliography{arr_main}

\appendix

\clearpage
\setcounter{page}{1}

\appendix
\setcounter{table}{0}   
\setcounter{figure}{0}
\setcounter{section}{0}
\setcounter{equation}{0}
\renewcommand{\thefigure}{\Alph{figure}} 
\renewcommand{\thetable}{\Alph{table}}

\addcontentsline{toc}{section}{Appendix} 

\section{Experiment Setting}
\label{appendix: Exp}
\subsection{Dataset}
\label{appendix: dataset}
\begin{table*}[t] 
\centering 
\caption{\textbf{Detailed breakdown of training data statistics.} The dataset is categorized into three distinct tasks: Caption, Open-Ended, and Multiple-Choice. For each task, we specify the number of samples and the provenance of the source datasets used to construct the SFT corpus.}
\label{tab:traing_datasets} 
\begin{tabular}{cccc}
\toprule 
\textbf{Modality} & \textbf{Task} & \textbf{Samples} & \textbf{Dataset} \\
\midrule 
\multirow{6}{*}{Video-Text} & \multirow{2}{*}{Caption} & \multirow{2}{*}{2k} & VidOR, YouCook2, Charades, ActivityNet, Kinetics-700\\
& & & Sthsth2, Ego4D, InternVid-10M, HD-VILA-100M, VIDAL \\
\cmidrule(lr){2-4}

& \multirow{2}{*}{Open-Ended} & \multirow{2}{*}{4k} & VidOR, YouCook2, Charades, ActivityNet\\
& & & Ego4D, InternVid-10M, HD-VILA-100M, VIDAL \\
\cmidrule(lr){2-4}

 & \multirow{2}{*}{Multiple-Choice} & \multirow{2}{*}{14k} & VidOR, YouCook2, Charades, ActivityNet\\
& & & Ego4D, InternVid-10M, HD-VILA-100M, VIDAL \\

\bottomrule 
\end{tabular}
\end{table*}
 For video-language training, we utilize large-scale video-text pairs sourced from several publicly accessible databases. The video training data is a subset of LLaVA-Video-178K, which includes VidOR, YouCook2, Charades, ActivityNet, Kinetics-700 Sthsth2, Ego4D, InternVid-10M, HD-VILA-100M, VIDAL. We randomly sampled 20,000 videos, including three different tasks: multiple-choice, caption and open-ended question. All the training data is demonstrated in Table~\ref{tab:traing_datasets}.

\subsection{Experiment Design}
 To thoroughly investigate the impact of Video-SFT on the visual capabilities of MLLMs, we design three experiments targeting model architecture, model scale, and training frame counts, respectively. The detailed configurations for these experiments are illustrated in Table~\ref{tab:my_experimental_setup}.

 \begin{figure*}[h]
  \centering
  \includegraphics[width=\linewidth]{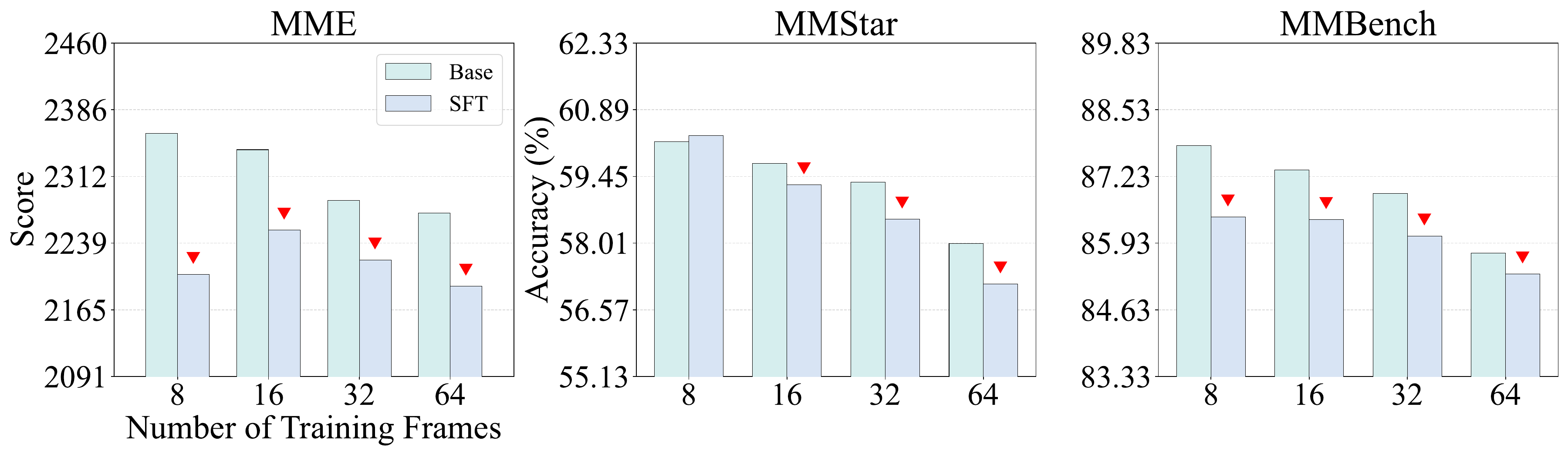}
    \caption{\textbf{Comparison of Qwen2.5-VL-7B model's performance on MME, MMStar and MMBench before and after 8, 16, 32 and 64 frames of Video-SFT, where the number of input images is the same as the number of frames used in Video-SFT.}  \textcolor{red}{$\blacktriangledown$} indicates performance degradation after Video-SFT. \textbf{Key Insight:} (1) SFT models generally lag behind Base models for any given number of frame. (2) A greater number of redundant image inputs hurt visual performance for MLLMs, justifying our \textbf{Hybrid-Frame Strategy}.}
  \label{fig:s_n_images}
\end{figure*}


 For the hybrid frame strategy, we employ Qwen3-VL-8B to evaluate a total of 74,500 QA pairs, aiming to adaptively determine the optimal number of frames required to resolve each instruction.  Utilizing the prompting strategy illustrated in Figure~\ref{fig:s_hybrid_frame_prompt}, and comprehensively evaluating aspects such as event duration, motion continuity, causality, object interaction, and fine-grained visual attributes, we obtain a frame distribution: 57,604 samples at 8 frames, 11,394 at 16 frames, 5,365 at 32 frames, and 137 at 64 frames, with an average of approximately 11 frames.

\section{Detailed Experimental Results}
\label{app:detailed experimental results}

\subsection{Detailed Results of Different Visual Tasks}
\label{app:detailed results of different visual tasks}
\begin{figure*}[tbp]
  \centering
  \includegraphics[width=\linewidth]    {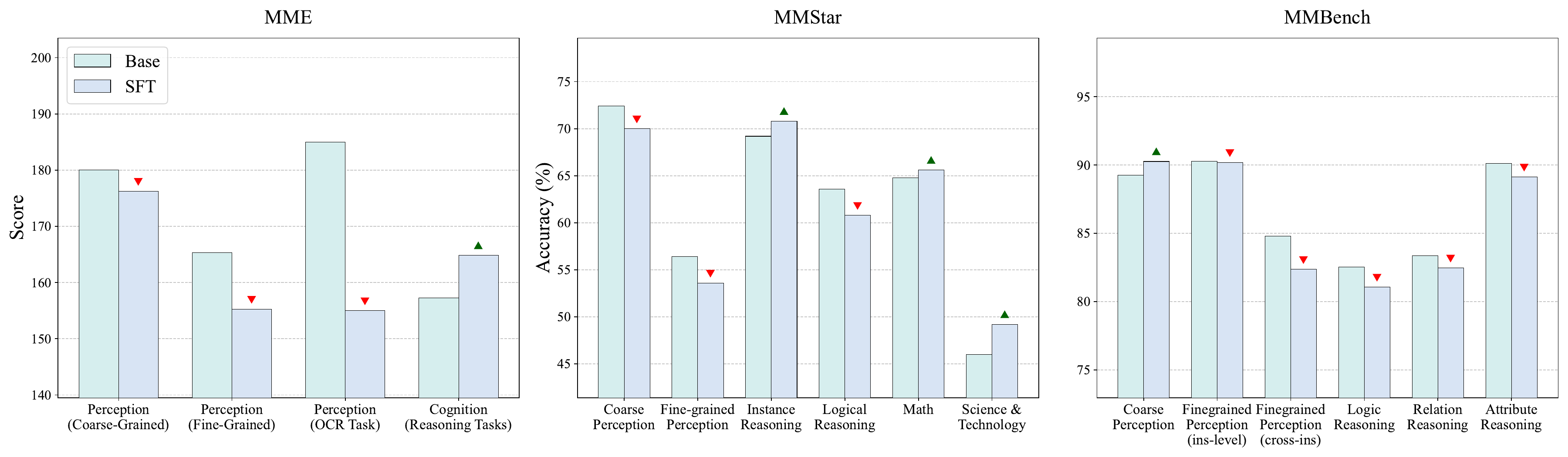}

\caption{\textbf{Performance impact of Video-SFT on Qwen2.5-VL-7B across Fine-Grained Perception, General Understanding, and Visual Reasoning tasks on MME, MMStar, and MMBench.} Each column corresponds to a specific capability dimension in image understanding. \textcolor{red}{$\blacktriangledown$} indicates performance degradation after Video-SFT, while \textcolor{darkgreen}{$\blacktriangle$} denotes improvement. The results reveal a non-uniform influence of Video-SFT across visual tasks, with Fine-Grained Perception exhibiting the most pronounced performance drop.
}
  
  \label{fig:3_subTask}
\end{figure*}

Figure~\ref{fig:3_subTask} illustrates the performance variations of Qwen2.5-VL-7B across different subtask categories within MME, MMStar, and MMBench.  In this section, we provide a more comprehensive set of experimental results, detailing the performance of three models, namely LLaVA-1.5-7B, LLaVA-NeXT-Video-7B, and Qwen2.5-VL-7B, before and after Video-SFT on the aforementioned benchmarks, with a breakdown by individual subtask. Tables~\ref{tab:s_qwen2.5_vl_mme_details},~\ref{tab:s_qwen2.5_vl_mmstar_details} and~\ref{tab:s_qwen2.5_mmbench_details} respectively present the experimental results on MME, MMStar and MMBench.

\begin{table*}[!ht] 
\centering 
\caption{\textbf{Experimental setup for assessing Video-SFT's effect on the visual capability of MLLMs.} We designed experiments from three aspects: model size, model architecture, and training frame count to explore the influence of Video-SFT on the visual capability of MLLMs.}
\label{tab:my_experimental_setup} 

\definecolor{lightgray}{gray}{0.9}

\resizebox{\linewidth}{!}{
\begin{tabular}{cccccc}
\toprule 
\multirow{2}{*}{\textbf{Task}} & \multirow{2}{*}{\textbf{Start Model}} & \multirow{2}{*}{\textbf{Training Frames}} & \multirow{2}{*}{\textbf{Eval Frames}} & \multicolumn{2}{c}{\textbf{Benchmark}} \\
\cmidrule(lr){5-6}
& & & & \textbf{Image} & \textbf{Video} \\
\midrule 

\multirow{4}{*}{Model Architecture} & \multirow{4}{*}{\begin{tabular}[c]{@{}c@{}}LLaVA-1.5-7B\\ LLaVA-NeXT-Video-7B\\ Qwen2.5-VL-7B\end{tabular}} & \multirow{4}{*}{8} & \multirow{4}{*}{8} & MME & Video-MME \\
& & & & MMStar & Video-MMMU \\
& & & & MMBench & MVBench \\
& & & & POPE & TempCompass \\

\midrule 

\multirow{4}{*}{Model Size} & \multirow{4}{*}{Qwen2.5-VL-3B/7B/32B/72B} & \multirow{4}{*}{8} & \multirow{4}{*}{8} & MME & Video-MME \\
& & & & MMStar & Video-MMMU \\
& & & & MMBench & MVBench \\
& & & & POPE & TempCompass \\

\midrule 

\multirow{3}{*}{Training Frames} & \multirow{3}{*}{\begin{tabular}[c]{@{}c@{}}Qwen2.5-VL-7B\end{tabular}} & \multirow{3}{*}{\begin{tabular}[c]{@{}c@{}}8/16/32/64\\ Hybrid-Frame\end{tabular}} & \multirow{3}{*}{\begin{tabular}[c]{@{}c@{}}8/16/32/64\\ Hybrid-Frame\end{tabular}} & \multirow{3}{*}{\begin{tabular}[c]{@{}c@{}}MMStar\\ POPE\end{tabular}} & \multirow{3}{*}{\begin{tabular}[c]{@{}c@{}}Video-MME\\ Video-MMMU\\ MVBench\end{tabular}} \\
& & & & & \\
& & & & & \\

\bottomrule 
\end{tabular}}
\vspace{-10pt}
\end{table*}

A consistent trend observed across all evaluated models is a degradation in static image perception capabilities following Video-SFT.    As shown by the MME and MMStar results, fine-grained tasks suffer the most significant performance drops.  Specifically, tasks requiring high spatial resolution, such as attribute and celebrity recognition, exhibit the sharpest decline.   For example, in the MME benchmark, the celebrity recognition score for LLaVA-1.5-7B dropped by 80.59 points, LLaVA-Next-Video dropped by 52.35 points and Qwen2.5-VL-2.5-7B dropped by 54.71 points.   Attribute recognition shows degradation across all three models in MMStar and MMBench.

Contrary to the significant decline in perception, cognitive reasoning capabilities demonstrate certain robustness. In several instances, Video-SFT appears to preserve or even enhance logical deduction capacity. On the MME benchmark, Qwen2.5-VL-7B showed improvements in numerical calculation (+7.50), code reasoning (+7.50), and text translation (+7.50) after Video-SFT. Similarly, on MMStar, the Math and Science \& Technology dimensions remained stable or improved slightly after Video-SFT. This suggests that while the perception  capacity may be damaged due to Video-SFT, the reasoning capacity  may benefit from the causal and sequential logic in video data, or at least remains stable.

\subsection{Impact of Input Differences during Training and Inference on Static Visual Capability}
\label{app:impact of input differences during training and inference on static visual capability}
In the Section~\ref{sec:4.1 train frame}, we evaluated image benchmarks using a single input image. To verify whether the performance degradation observed in Video-SFT models stems from a discrepancy between the training modality (multi-frame videos) and the inference modality (single images).We conducted a controlled experiment where a single static image is copied multiple times to simulate video input, with the number of copies being the same as the number of frames used during Video-SFT.   This setup ensures that the context length and input structure match those of Video-SFT, while the semantic content remains static.   We compared the performance of the Qwen2.5-VL-7B model in MME, MMStar and MMBench before and after 8, 16, 32 and 64 frames of Video-SFT. The results are shown in Figure~\ref{fig:s_n_images}

Degradation persists under identical input modalities. For any given number of frames, the Video-SFT model consistently underperforms the Base model.  This confirms that the degradation of the model's static visual ability is mainly caused by Video-SFT, rather than the format differences between inference and training inputs.
We observe a consistent trend where increasing the number of replicated frames leads to performance drops.   For example, in Base model, as the number of input images increased from 8 to 64, the score of MME dropped from 2360 to 2272, the accuracy of MMStar decreased from 60.20 to 58.00, and the accuracy of MMBench declined from 87.83 to 85.74.   This indicates that simply padding the same static information into long temporal sequences introduces redundancy and damages the visual ability of the model.    This finding provides strong support for our proposed hybrid frame strategy, which adaptively reduces frame counts for redundant content to avoid degradation caused by temporal trap.

\subsection{Impact of Training Strategies under Unified Video Inputs}
\label{app:impact of training strategies under unified video inputs}
We compare performance of Qwen2.5-VL-7B under different Video-SFT strategies across multiple frame settings in Table~\ref{tab:total_model_frame}. The number of frames used for model inference is the same as that used for Video-SFT.  In real-world deployment scenarios, MLLMs are often constrained by latency and computational budgets, necessitating inference with fewer frames than were available during training, such as 8 frames.   To evaluate the robustness of different Video-SFT strategies under this constraint, we conducted an experiment where  Qwen2.5-VL-7B trained with different frame counts (8, 16, 32, 64, and our Hybrid strategy) were all evaluated using a fixed input of 8 frames across video benchmarks.
The experimental results are shown in Table~\ref{tab:s_all_of_8_frames_video}. Compared with the basic model, the performance of each Video-SFT model adopting different frame rate strategies has been improved.

In the MVBench dataset, the performance of the models trained with 16, 32, and 64 frames all decreased significantly compared to those trained with 8 frames. However, our hybrid frame strategy maintained an accuracy rate of 63.94\%, which was much higher than that of the models trained with 16, 32, and 64 frames. Meanwhile, in the Video-MME and Video-MMMU benchmarks, the hybrid frame strategy also shows high performance, which indicates that the hybrid frame strategy has strong robustness.

\begin{table}[h]
    \centering
    \small
    \setlength{\tabcolsep}{5pt}
    \caption{\textbf{Performance comparison of Qwen2.5-VL-7B under different Video-SFT strategies across multiple frame settings (8, 16, 32, 64, and our Hybrid Strategy) on Video-MME, Video-MMMU, and MVBench.} All models are evaluated using a fixed input of 8 frames. Numbers in \textcolor{darkgreen}{$\uparrow$} indicate improvements relative to the base model.}
    \label{tab:s_all_of_8_frames_video}
    \resizebox{\columnwidth}{!}{
    \begin{tabular}{lcccc}
        \toprule
        \textbf{Setting} & \textbf{Video-MME} & \textbf{Video-MMMU} & \textbf{MVBench} \\
        \midrule
        \multicolumn{4}{l}{\textbf{Inference with 8 Frames}} \\
        \midrule
        Qwen2.5-VL-7B (Base) & 51.19 & 44.22 & 62.68 \\
        \midrule
        + Video-SFT-8F & 54.41 {\color{darkgreen}(↑3.22)} & 47.56 {\color{darkgreen}(↑3.34)} & 64.62 {\color{darkgreen}(↑1.94)} \\
        + Video-SFT-16F & 54.52 {\color{darkgreen}(↑3.33)} & 48.11 {\color{darkgreen}(↑3.89)} & 63.05 {\color{darkgreen}(↑0.37)} \\
        + Video-SFT-32F & 55.67 {\color{darkgreen}(↑4.48)} & 47.44 {\color{darkgreen}(↑3.22)} & 63.17 {\color{darkgreen}(↑0.49)} \\
        + Video-SFT-64F & 55.26 {\color{darkgreen}(↑4.07)} & 47.44 {\color{darkgreen}(↑3.22)} & 63.05 {\color{darkgreen}(↑0.37)} \\
        \midrule
        \rowcolor[RGB]{245,248,255}
        + Hybrid-Frame (Ours) & 54.96 {\color{darkgreen}(↑3.77)} & 47.22 {\color{darkgreen}(↑3.00)} & 63.94 {\color{darkgreen}(↑1.26)} \\
        \bottomrule
    \end{tabular}
    }
\end{table}


\section{Discussion}
\label{app:discussion}

\subsection{Hybrid-Frame Strategy}

\newcolumntype{P}{>{\centering\arraybackslash}p{2.6cm}}

\begin{table*}[htbp]
\caption{\textbf{Performance comparison of Qwen2.5-VL-7B, LLaVA-Next Video-7B and LLaVA-1.5-7B before and after Video-SFT in different tasks on the MME Benchmark.}   Numbers in \textcolor{darkgreen}{$\uparrow$} indicate improvements, while numbers in \textcolor{red}{$\downarrow$} indicate degradations relative to the base model.}
\centering
\resizebox{\textwidth}{!}{%
\begin{tabular}{lPPPPPP}
\toprule
\multirow{3}{*}{\textbf{Task Categories}} & \multicolumn{6}{c}{\textbf{Models}} \\
\cmidrule(lr){2-7}
 & \multicolumn{2}{c}{\textbf{Qwen2.5-VL-7B}} & \multicolumn{2}{c}{\textbf{LLaVA-Next-Video-7B}} & \multicolumn{2}{c}{\textbf{LLaVA-1.5-7B}} \\
 & \textbf{Base} & \textbf{SFT} & \textbf{Base} & \textbf{SFT} & \textbf{Base} & \textbf{SFT} \\
\midrule

\multicolumn{7}{c}{\cellcolor{gray!10}\textbf{Perception (Coarse-grained)}} \\
\midrule
\textbf{Existence} & 195.00 & 190.00 {\color{red}(↓5.00)} & 200.00 & 190.00 {\color{red}(↓10.00)} & 195.00 & 135.00 {\color{red}(↓60.00)} \\
\textbf{Count} & 165.00 & 170.00 {\color{darkgreen}(↑5.00)} & 135.00 & 153.33 {\color{darkgreen}(↑18.33)} & 148.33 & 105.00 {\color{red}(↓43.33)} \\
\textbf{Position} & 165.00 & 160.00 {\color{red}(↓5.00)} & 121.67 & 130.00 {\color{darkgreen}(↑8.33)} & 128.33 & 51.67 {\color{red}(↓76.66)} \\
\textbf{Color} & 195.00 & 185.00 {\color{red}(↓10.00)} & 150.00 & 145.00 {\color{red}(↓5.00)} & 150.00 & 101.67 {\color{red}(↓48.33)} \\

\midrule
\multicolumn{7}{c}{\cellcolor{gray!10}\textbf{Perception (Fine-grained)}} \\
\midrule
\textbf{Poster} & 173.13 & 183.33 {\color{darkgreen}(↑10.20)} & 141.50 & 145.92 {\color{darkgreen}(↑4.42)} & 142.52 & 95.58 {\color{red}(↓46.94)} \\
\textbf{Celebrity} & 165.00 & 110.29 {\color{red}(↓54.71)} & 136.47 & 84.12 {\color{red}(↓52.35)} & 135.88 & 55.29 {\color{red}(↓80.59)} \\
\textbf{Scene} & 155.50 & 161.50 {\color{darkgreen}(↑6.00)} & 158.00 & 152.50 {\color{red}(↓5.50)} & 156.50 & 129.00 {\color{red}(↓27.50)} \\
\textbf{Landmark} & 184.75 & 183.50 {\color{red}(↓1.25)} & 154.75 & 158.50 {\color{darkgreen}(↑3.75)} & 157.00 & 115.00 {\color{red}(↓42.00)} \\
\textbf{Artwork} & 148.00 & 137.75 {\color{red}(↓10.25)} & 110.25 & 109.50 {\color{red}(↓0.75)} & 116.50 & 73.25 {\color{red}(↓43.25)} \\

\midrule
\multicolumn{7}{c}{\cellcolor{gray!10}\textbf{Perception (OCR)}} \\
\midrule
\textbf{OCR} & 185.00 & 155.00 {\color{red}(↓30.00)} & 140.00 & 87.50 {\color{red}(↓52.50)} & 147.50 & 50.00 {\color{red}(↓97.50)} \\

\midrule
\multicolumn{7}{c}{\cellcolor{gray!10}\textbf{Cognition (Reasoning)}} \\
\midrule
\textbf{Commonsense Reasoning} & 141.43 & 149.29 {\color{darkgreen}(↑7.86)} & 117.86 & 115.71 {\color{red}(↓2.15)} & 112.14 & 76.43 {\color{red}(↓35.71)} \\
\textbf{Numerical Calculation} & 140.00 & 147.50 {\color{darkgreen}(↑7.50)} & 55.00 & 40.00 {\color{red}(↓15.00)} & 47.50 & 50.00 {\color{darkgreen}(↑2.50)} \\
\textbf{Text Translation} & 185.00 & 192.50 {\color{darkgreen}(↑7.50)} & 97.50 & 65.00 {\color{red}(↓32.50)} & 55.00 & 50.00 {\color{red}(↓5.00)} \\
\textbf{Code Reasoning} & 162.50 & 170.00 {\color{darkgreen}(↑7.50)} & 57.50 & 42.50 {\color{red}(↓15.00)} & 75.00 & 50.00 {\color{red}(↓25.00)} \\

\bottomrule
\end{tabular}%
}
\label{tab:s_qwen2.5_vl_mme_details}
\end{table*}
Although our current experiment demonstrates the efficacy of hybrid frame strategy, the current approach relies on a predefined discrete set of frame counts (8, 16, 32, 64) and operates solely on textual instruction. Moving forward, we envision a more robust and effective decision-making framework.

\textbf{From text to visual content.}
A limitation of the current strategy is its blindness to the actual visual dynamics of the video. A complex textual query might correspond to a static scene or a high-speed sequence, necessitating vastly different temporal resolutions. To bridge this gap, future iterations should incorporate visual-content awareness into the decision loop.      We can use visual encoders to compute inter-frame feature similarity. By analyzing the visual redundancy and motion magnitude, the system can determine whether high frame rates are genuinely required. For example, videos with high inter-frame similarity scores could be sampled sparsely regardless of the textual prompt complexity, thereby eliminating redundant computation.

\textbf{Multi-model decision-making.}
Relying on a single model for decision introduces potential biases or hallucinations, particularly for ambiguous instructions. To enhance robustness and accuracy, we can use a multi-model consensus mechanism. By deploying  diverse models to independently assess, we can aggregate their outputs via weighted voting.   This "multi-model decision-making" approach ensures that the selected frame count reflects a generalized understanding of the task requirements, minimizing the risk of error due to the failure of a single model.

\textbf{Continuous and adaptive granularity.}
The current method selects from fixed discrete frame counts and has many limitations.    By combining the aforementioned visual redundancy analysis with the multi-model consensus, we can regress a precise, arbitrary frame count that strictly aligns with the spatio-temporal information of the input. It will further improve efficiency and accuracy.

\subsection{General-Purpose Multimodal Models}

Recent advancements in MLLMs have increasingly emphasized developing general-purpose multimodal models capable of seamlessly processing text, images, video, and audio within a unified framework.    The unification of the two visual modalities—images and videos—is a crucial step.

In this work, we conduct a systematic analysis of MLLMs under Video-SFT.    We explore whether current MLLMs can jointly benefit from Video-SFT across both static and temporal visual tasks.    Our experiments reveal that current efforts aimed at enhancing the visual understanding of MLLMs often encounter conflicts between image and video modalities.   

Improvements in video reasoning fail to transfer—or even degrade—image understanding.    This reveals the temporal trap in visual modality conflict, which might be because temporal supervision undermines spatial generalization.    These findings indicate that existing unified training pipelines have not yet achieved true modality synergy.
Our Hybrid-Frame Strategy represents a preliminary attempt at the unification of image and video processing, providing ideas and experience for the future construction of a unified visual or even a universal modal training paradigm.
\newcolumntype{P}{>{\centering\arraybackslash}p{2.4cm}}

\begin{table*}[htbp]
\caption{\textbf{Performance comparison of Qwen2.5-VL-7B, LLaVA-Next Video-7B and LLaVA-1.5-7B before and after Video-SFT in different tasks on the MMStar Benchmark.}   Numbers in \textcolor{darkgreen}{$\uparrow$} indicate improvements, while numbers in \textcolor{red}{$\downarrow$} indicate degradations relative to the base model.}
\centering
\resizebox{\textwidth}{!}{%
\begin{tabular}{lPPPPPP}
\toprule
\multirow{3}{*}{\textbf{Task Capabilities}} & \multicolumn{6}{c}{\textbf{Models}} \\
\cmidrule(lr){2-7}
 & \multicolumn{2}{c}{\textbf{Qwen2.5-VL-7B}} & \multicolumn{2}{c}{\textbf{LLaVA-Next-Video-7B}} & \multicolumn{2}{c}{\textbf{LLaVA-1.5-7B}} \\
 & \textbf{Base} & \textbf{SFT} & \textbf{Base} & \textbf{SFT} & \textbf{Base} & \textbf{SFT} \\
\midrule

\multicolumn{7}{c}{\cellcolor{gray!10}\textbf{Coarse Perception}} \\
\midrule
\textbf{Image Scene \& Topic} & 63.83 & 60.99 {\color{red}(↓2.84)} & 51.77 & 48.23 {\color{red}(↓3.54)} & 49.65 & 32.62 {\color{red}(↓17.03)} \\
\textbf{Image Emotion} & 70.97 & 67.74 {\color{red}(↓3.23)} & 80.65 & 67.74 {\color{red}(↓12.91)} & 80.65 & 54.84 {\color{red}(↓25.81)} \\
\textbf{Image Style \& Quality} & 88.46 & 87.18 {\color{red}(↓1.28)} & 69.23 & 75.64 {\color{darkgreen}(↑6.41)} & 69.23 & 38.46 {\color{red}(↓30.77)} \\

\midrule
\multicolumn{7}{c}{\cellcolor{gray!10}\textbf{Fine-grained Perception}} \\
\midrule
\textbf{Attribute \& Celebrity} & 57.63 & 53.39 {\color{red}(↓4.24)} & 34.75 & 27.97 {\color{red}(↓6.78)} & 28.81 & 25.42 {\color{red}(↓3.39)} \\
\textbf{Object Counting} & 48.91 & 47.83 {\color{red}(↓1.08)} & 28.26 & 33.70 {\color{darkgreen}(↑5.44)} & 22.83 & 28.26 {\color{darkgreen}(↑5.43)} \\
\textbf{Object Location} & 70.00 & 67.50 {\color{red}(↓2.50)} & 45.00 & 27.50 {\color{red}(↓17.50)} & 20.00 & 32.50 {\color{darkgreen}(↑12.50)} \\

\midrule
\multicolumn{7}{c}{\cellcolor{gray!10}\textbf{Instance Reasoning}} \\
\midrule
\textbf{Single-Instance Attribute Reasoning} & 68.69 & 68.69  & 57.58 & 51.52 {\color{red}(↓6.06)} & 53.54 & 42.42 {\color{red}(↓11.12)} \\
\textbf{Cross-Instance Attribute Comparison} & 67.42 & 68.54 {\color{darkgreen}(↑1.12)} & 32.58 & 31.46 {\color{red}(↓1.12)} & 26.97 & 20.22 {\color{red}(↓6.75)} \\
\textbf{Cross-Instance Relation Reasoning} & 72.58 & 77.42 {\color{darkgreen}(↑4.84)} & 33.87 & 24.19 {\color{red}(↓9.68)} & 29.03 & 16.13 {\color{red}(↓12.90)} \\

\midrule
\multicolumn{7}{c}{\cellcolor{gray!10}\textbf{Logical Reasoning}} \\
\midrule
\textbf{Diagram Reasoning} & 68.18 & 68.18  & 24.55 & 21.82 {\color{red}(↓2.73)} & 21.82 & 31.82 {\color{darkgreen}(↑10.00)} \\
\textbf{Common Reasoning} & 63.37 & 56.44 {\color{red}(↓6.93)} & 37.62 & 37.62  & 31.68 & 23.76 {\color{red}(↓7.92)} \\
\textbf{Code \& Sequence Reasoning} & 51.28 & 51.28  & 41.03 & 33.33 {\color{red}(↓7.70)} & 30.77 & 30.77  \\

\midrule
\multicolumn{7}{c}{\cellcolor{gray!10}\textbf{Mathematics}} \\
\midrule
\textbf{Numeric Commonsense \& Calculation} & 62.50 & 62.50  & 18.75 & 37.50 {\color{darkgreen}(↑18.75)} & 27.08 & 31.25 {\color{darkgreen}(↑4.17)} \\
\textbf{Statistical Analysis} & 60.47 & 62.79 {\color{darkgreen}(↑2.32)} & 36.05 & 31.40 {\color{red}(↓4.65)} & 26.74 & 30.23 {\color{darkgreen}(↑3.49)} \\
\textbf{Geometry} & 68.97 & 68.97  & 23.28 & 21.55 {\color{red}(↓1.73)} & 23.28 & 24.14 {\color{darkgreen}(↑0.86)} \\

\midrule
\multicolumn{7}{c}{\cellcolor{gray!10}\textbf{Science \& Technology}} \\
\midrule
\textbf{Natural Science} & 47.26 & 49.32 {\color{darkgreen}(↑2.06)} & 23.29 & 17.81 {\color{red}(↓5.48)} & 18.49 & 22.60 {\color{darkgreen}(↑4.11)} \\
\textbf{Engineering} & 41.30 & 45.65 {\color{darkgreen}(↑4.35)} & 17.39 & 17.39  & 28.26 & 26.09 {\color{red}(↓2.17)} \\
\textbf{Geography \& Earth Science} & 46.55 & 51.72 {\color{darkgreen}(↑5.17)} & 50.00 & 32.76 {\color{red}(↓17.24)} & 27.59 & 27.59  \\

\bottomrule
\end{tabular}%
}
\label{tab:s_qwen2.5_vl_mmstar_details}
\end{table*}
\newcolumntype{P}{>{\centering\arraybackslash}p{2.6cm}}

\begin{table*}[h]
\caption{\textbf{Performance comparison of Qwen2.5-VL-7B, LLaVA-Next Video-7B and LLaVA-1.5-7B before and after Video-SFT in different tasks on the MMBench Benchmark.}   Numbers in \textcolor{darkgreen}{$\uparrow$} indicate improvements, while numbers in \textcolor{red}{$\downarrow$} indicate degradations relative to the base model.}
\centering
\resizebox{\textwidth}{!}{
\begin{tabular}{lPPPPPP}
\toprule
\multirow{3}{*}{\textbf{Task Categories}} & \multicolumn{6}{c}{\textbf{Models}} \\
\cmidrule(lr){2-7}
 & \multicolumn{2}{c}{\textbf{Qwen2.5-VL-7B}} & \multicolumn{2}{c}{\textbf{LLaVA-Next-Video-7B}} & \multicolumn{2}{c}{\textbf{LLaVA-1.5-7B}} \\
 & \textbf{Base} & \textbf{SFT} & \textbf{Base} & \textbf{SFT} & \textbf{Base} & \textbf{SFT} \\
\midrule

\multicolumn{7}{c}{\cellcolor{gray!10}\textbf{Coarse Perception}} \\
\midrule
\textbf{Image Scene} & 98.28 & 98.28  & 79.12 & 73.96 {\color{red}(↓5.16)} & 78.38 & 47.17 {\color{red}(↓31.21)} \\
\textbf{Image Emotion} & 83.00 & 84.50 {\color{darkgreen}(↑1.50)} & 54.50 & 70.00 {\color{darkgreen}(↑15.50)} & 74.00 & 43.50 {\color{red}(↓30.50)} \\
\textbf{Image Quality} & 58.67 & 62.67 {\color{darkgreen}(↑4.00)} & 36.67 & 40.67 {\color{darkgreen}(↑4.00)} & 37.33 & 34.67 {\color{red}(↓2.66)} \\
\textbf{Image Topic} & 93.57 & 94.29 {\color{darkgreen}(↑0.72)} & 76.43 & 68.57 {\color{red}(↓7.86)} & 74.29 & 50.71 {\color{red}(↓23.58)} \\
\textbf{Image Style} & 96.70 & 97.17 {\color{darkgreen}(↑0.47)} & 61.79 & 58.49 {\color{red}(↓3.30)} & 58.02 & 28.77 {\color{red}(↓29.25)} \\

\midrule
\multicolumn{7}{c}{\cellcolor{gray!10}\textbf{Fine-grained Perception (Single-Instance)}} \\
\midrule
\textbf{Celebrity Recognition} & 96.97 & 96.72 {\color{red}(↓0.25)} & 64.14 & 69.19 {\color{darkgreen}(↑5.05)} & 68.94 & 27.53 {\color{red}(↓41.41)} \\
\textbf{Object Localization} & 73.33 & 73.33  & 48.57 & 40.63 {\color{red}(↓7.94)} & 40.32 & 27.94 {\color{red}(↓12.38)} \\
\textbf{Attribute Recognition} & 97.73 & 97.35 {\color{red}(↓0.38)} & 68.18 & 62.88 {\color{red}(↓5.30)} & 70.45 & 38.26 {\color{red}(↓32.19)} \\
\textbf{OCR} & 94.87 & 95.51 {\color{darkgreen}(↑0.64)} & 54.49 & 62.82 {\color{darkgreen}(↑8.33)} & 54.49 & 51.92 {\color{red}(↓2.57)} \\

\midrule
\multicolumn{7}{c}{\cellcolor{gray!10}\textbf{Fine-grained Perception (Cross-Instance)}} \\
\midrule
\textbf{Action Recognition} & 94.88 & 94.42 {\color{red}(↓0.46)} & 82.33 & 68.37 {\color{red}(↓13.96)} & 73.49 & 45.58 {\color{red}(↓27.91)} \\
\textbf{Attribute Comparison} & 94.33 & 90.07 {\color{red}(↓4.26)} & 36.17 & 17.02 {\color{red}(↓19.15)} & 19.86 & 18.44 {\color{red}(↓1.42)} \\
\textbf{Spatial Relationship} & 64.97 & 61.58 {\color{red}(↓3.39)} & 32.77 & 33.90 {\color{darkgreen}(↑1.13)} & 31.07 & 32.20 {\color{darkgreen}(↑1.13)} \\

\midrule
\multicolumn{7}{c}{\cellcolor{gray!10}\textbf{Attribute Reasoning}} \\
\midrule
\textbf{Function Reasoning} & 94.08 & 92.76 {\color{red}(↓1.32)} & 62.50 & 61.18 {\color{red}(↓1.32)} & 65.13 & 50.33 {\color{red}(↓14.80)} \\
\textbf{Physical Property} & 77.17 & 76.71 {\color{red}(↓0.46)} & 42.92 & 51.14 {\color{darkgreen}(↑8.22)} & 55.25 & 39.73 {\color{red}(↓15.52)} \\
\textbf{Identity Reasoning} & 99.43 & 98.30 {\color{red}(↓1.13)} & 92.61 & 83.52 {\color{red}(↓9.09)} & 48.86 & 44.32 {\color{red}(↓4.54)} \\

\midrule
\multicolumn{7}{c}{\cellcolor{gray!10}\textbf{Relation Reasoning}} \\
\midrule
\textbf{Natural Relation} & 87.71 & 86.59 {\color{red}(↓1.12)} & 45.25 & 34.64 {\color{red}(↓10.61)} & 41.90 & 20.11 {\color{red}(↓21.79)} \\
\textbf{Physical Relation} & 60.64 & 58.51 {\color{red}(↓2.13)} & 44.68 & 41.49 {\color{red}(↓3.19)} & 41.49 & 37.23 {\color{red}(↓4.26)} \\
\textbf{Social Relation} & 91.28 & 91.28  & 58.72 & 63.95 {\color{darkgreen}(↑5.23)} & 69.19 & 30.81 {\color{red}(↓38.38)} \\

\midrule
\multicolumn{7}{c}{\cellcolor{gray!10}\textbf{Logic Reasoning}} \\
\midrule
\textbf{Future Prediction} & 70.00 & 63.85 {\color{red}(↓6.15)} & 38.46 & 20.77 {\color{red}(↓17.69)} & 38.46 & 33.08 {\color{red}(↓5.38)} \\
\textbf{Image-Text Understanding} & 88.30 & 89.01 {\color{darkgreen}(↑0.71)} & 41.49 & 41.84 {\color{darkgreen}(↑0.35)} & 37.23 & 32.27 {\color{red}(↓4.96)} \\

\bottomrule
\end{tabular}%
}
\label{tab:s_qwen2.5_mmbench_details}
\end{table*}

   \begin{figure*}[h]
\definecolor{promptblue}{RGB}{81, 149, 211}
\definecolor{promptblueck}{RGB}{232, 238, 247}

\lstdefinestyle{PythonStyle}{
    language=Python,
    basicstyle=\small\ttfamily,  
    breaklines=true,             
    frame=single,                
    backgroundcolor=\color{gray!10}, 
}

\centering 
\vspace{40pt}
\vspace*{\fill} 
\begin{tcolorbox}[
    colframe=promptblue,          
    colback=promptblueck,         
    title={\textbf{Hybrid-Frame Strategy Prompt}}, 
    fonttitle=\color{white}\bfseries, 
]

You are an expert Video Data Strategist. Your task is to perform a holistic assessment of a QA pair to estimate the optimal video frame count required to faithfully answer the instruction.
\\ \\
Objective:\\
You must infer the necessary temporal evidence and spatial granularity based on the text descriptions, determining the adaptive frame count 8, 16, 32, or 64.
\\ \\
Holistic Assessment Framework:\\
Analyze the input QA pair across the following five dimensions defined in your architecture:

1. Event Duration: Does the text describe a long-term state (low frames) or a fleeting, split-second occurrence (high frames)?\\
2. Motion Continuity: Is the action smooth and continuous, or does it involve rapid, disjointed changes?\\
3. Causal Relations: Does the answer require understanding a "cause and effect" sequence (e.g., "Why did the glass break?") which requires higher temporal density to see the trigger?\\
4. Object Interactions: Are there complex interactions between objects (e.g., handing something over, fighting) that require precise tracking?\\
5. Fine-Grained Visual Attributes: Does the instruction demand high spatial clarity for small details (e.g., reading text, checking eye color)?
\\ \\ 
Decision Logic:

8 Frames (Global/Static Baseline):
Criteria: Low temporal dependency. The answer relies on static scenes, global atmosphere, or objects with no significant interaction or motion.
Dimensions: Long duration, zero motion, no causal complexity.

16 Frames (Coarse Temporal Evidence):
Criteria: The instruction focuses on a clear, single action or distinct event.
Dimensions: Moderate duration, simple motion continuity, clear subject-object relation.

32 Frames (Dense Spatiotemporal Reasoning):
Criteria: Required for multi-step sequences, procedural descriptions, or complex object interactions.
Dimensions: Strong causal relations (sequence matters), multiple object interactions.

64 Frames (High-Fidelity Granularity):
Criteria: Critical for analyzing micro-events, high-speed motion, or finding "needle-in-a-haystack" visual attributes.
Dimensions: Fleeting duration (instant), high motion complexity, or extreme need for fine-grained visual attributes.
\\ \\
Input QA Pair:\\

\{qa\_string\}\\
\\
Output:\\
Based on the assessment of the dimensions above, output ONLY the single integer representing the optimal frame count 8, 16, 32, or 64.
\end{tcolorbox}

\caption{\textbf{The prompt used for Hybrid-Frame Strategy.} It comprehensively evaluates aspects such as event duration, motion continuity, causality, object interaction, and fine-grained visual attributes, aiming to adaptively determine the optimal number of frames required to resolve each instruction.}
\label{fig:s_hybrid_frame_prompt}

\vspace{40pt}
\end{figure*}

\label{sec:appendix}

\end{document}